\documentclass[11pt]{article}
\usepackage{coling2020}
\usepackage{times}
\usepackage{url}
\usepackage{latexsym}
\usepackage[pdftex]{graphicx}
\usepackage{subcaption}

\usepackage{float}
\usepackage{booktabs}
\usepackage{multirow}
\usepackage{array}
\usepackage{xcolor}
\usepackage[toc,page]{appendix}

\usepackage{footnote}
\makesavenoteenv{tabular}
\makesavenoteenv{table}

\title{Evaluating Cross-Lingual Transfer Learning Approaches in Multilingual Conversational Agent Models}

\author{Lizhen Tan \\
  Amazon\\
  {\tt ltn@amazon.com} \\\And
  Olga Golovneva \\
  Amazon\\
  {\tt olggol@amazon.com} \\}

\date{}

\begin{document}
\maketitle
\begin{abstract}
	With the recent explosion in popularity of voice assistant devices, there is a growing interest in making them available to user populations in additional countries and languages. However, to provide the highest accuracy and best performance for specific user populations, most existing voice assistant models are developed individually for each region or language, which requires linear investment of effort. In this paper, we propose a general multilingual model framework for Natural Language Understanding (NLU) models, which can help bootstrap new language models faster and reduce the amount of effort required to develop each language separately. We explore how different deep learning architectures affect multilingual NLU model performance. Our experimental results show that these multilingual models can reach same or better performance compared to monolingual models across language-specific test data while require less effort in creating features and model maintenance.
\end{abstract}

\section{Introduction}
\blfootnote{
    %
    %
    %
    %
 \hspace{-0.65cm}  
 This work is licensed under a Creative Commons
 Attribution 4.0 International License. \\
 License details:
 \url{http://creativecommons.org/licenses/by/4.0/}.
}
The recent surge in popularity of voice assistants, such as Google Home, Apple’s Siri, or Amazon’s Alexa resulted in interest in scaling these products to more regions and languages. This means that all the components supporting Spoken Language Understanding (SLU) in these devices, such as Automatic Speech Recognition (ASR), Natural Language Understanding (NLU), and Entity Resolution (ER) are facing the challenges of scaling the development and maintenance processes for multiple languages and dialects.

When a voice assistant is launched in a new locale, its underlying speech processing components are often developed
specifically for the targeted country, marketplace, and the main language variant of that country. Many people assume that if a device ``understands'' and ``speaks'' in a specific language, for example English, it should be able to work equally well for any English-speaking country, but this is a misunderstanding. For instance, if a speaker of UK English asks a device trained on data collected in the United States ``\textit{tell me a famous football player}'', it is highly unlikely that this device will provide the user's desired answer, since \textit{football} means different things in the US and UK cultures. As a result, developers need to take into account not only the language or dialectal differences, but also local culture, to provide the right information in the right language setup. An increase in the number of target marketplaces often means a linear increase in effort needed to develop and maintain such locale-specific models.

NLU models, which classify the user’s intent and extract any significant entities from the user’s utterance, face the same challenge of maintaining high accuracy while being able to accommodate multiple dialects or language content. The major tasks in NLU are intent classification and slot filling. Intent classification is a task to predict what action the user intends the voice assistant to take. Slot filling is a task to identify the specific semantic arguments for the intention. For example, if the user’s request is to ``\textit{play Poker Face by Lady Gaga}'', the user’s intention will be ``\textit{play music}'', while in order to fulfill this command with specified details, the system needs to capture the slots for \{\textit{song name = Poker Face}\}, and \{\textit{artist name = Lady Gaga}\}. These tasks are called intent classification (IC) and named entity recognition (NER), respectively.

One common approach is to use a max-entropy (MaxEnt) classification model for the IC task and a conditional random fields (CRF) model for the NER task. Following the advent of deep learning techniques in related fields, such as computer vision and natural language processing, deep learning is becoming more popular in NLU as well. Some of the recent multilingual approaches to NLU include, for example, the Convolutional Neural Network (CNN) model for sentence classification \cite{kim-2014-convolutional}, or the Long Short-Term Memory (LSTM) model for NER prediction \cite{lample-etal-2016-neural,DBLP:journals/corr/KurataXZY16}. In the deep neural network architecture, the aforementioned NLU tasks can be combined into a single multi-task classification model. An increasing number of experiments also focus on multilingual setups, especially in the field of machine translation, where the task is to translate input from one language to another \cite{DBLP:journals/corr/JohnsonSLKWCTVW16}.

One recent thread of multilingual research centers around learning multilingual word representation. Multilingual word embeddings in the shared cross-lingual vector space have one main property: words from different languages but with similar meaning must be geometrically close. This property allows for transfer learning from one language to another in various multilingual tasks, such as dependency parsing \cite{DBLP:journals/corr/abs-1904-02099,wang-etal-2019-cross} or classification and NER \cite{DBLP:journals/corr/abs-1901-07291,pires-etal-2019-multilingual}. A number of model architectures have been proposed to pre-train multilingual word representations, such as leveraging large-scaled LSTM networks trained on monolingual corpora and adversarial setup for space alignment \cite{1,DBLP:journals/corr/abs-1711-00043}, or transformers trained on multilingual corpora as a single language model \cite{DBLP:journals/corr/abs-1901-07291}.

Although some of these models can be used to solve IC and NER tasks by appending corresponding decoders to generate final predictions, it is not straightforward to use them in production environments due to latency and memory constrains. A different way of benefitting from larger models could be to use them for transfer learning to smaller-size models to improve their performance by initializing some parts of the model with close-to-optimal rather than random weights. In this paper, we extend the multi-task approach studied in \cite{DBLP:journals/corr/abs-1904-01825} to a general multilingual model for IC and NER tasks, based on deep learning techniques, such as a bidirectional Long Short-Term Memory (biLSTM) CRF sequence labeling model for NER along with a multilayer perceptron (MLP) for IC.

We also explore multilingual transfer learning and its benefits to our setup. Transfer learning is widely adapted for zero-shot or few-shot setups, and was explored in some multilingual NLP studies \cite{mcdonald-etal-2011-multi,naseem-etal-2012-selective,DBLP:journals/corr/abs-1810-03552}, and also has been used in multi-task IC-NER models \cite{DBLP:journals/corr/abs-1904-01825},  yet to the best of our knowledge, there is no study applying transfer learning for data-rich target languages in a multilingual setup. In our experiment, we apply few-shot transfer learning from data-rich languages to a language with a smaller amout of training data. In additon, we also apply  transfer learning to mimic the situation of expanding the model ability to same-level-resource language with known context from another high-resource language(s), such that the new multilingual model will ``inherit'' context information from its ancestors. We investigate these approaches to transfer learning and their effects on model performance. We show that transfer learning can improve NLU model performance even in data-rich conditions.

\section{Multilingual model architecture for DC, IC, and NER tasks}

Single-language models are usually trained with data in a specific language, yet the model architecture is shared among different model instances (one per language). It follows that using the same model architecture, we should be able to train a generalized multilingual model which is fed by data from multiple languages.

\begin{figure}[ht]
	\centering
	\includegraphics[width=0.45\linewidth]{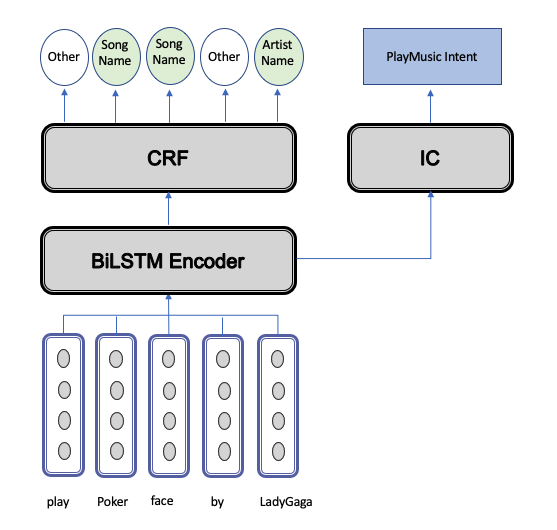}
	\caption{ Multi-task model for IC and NER prediction. Tokens are first embedded into word vectors, then fed in to a biLSTM encoder, whose output is used in both CRF and MLP for NER and IC, respectively.}\label{fig:1}
\end{figure}

NLU model is first trained to recognize the utterance domain, such as Music, Weather, Notifications, and then it is trained to perform domain-specific IC and NER tasks. In our experiments, the domain classification (DC) model is a MaxEnt logistic regression model. Despite the relatively simple architecture, this model shows good performance on DC task in data-rich conditions. For IC and NER tasks, we build a multi-task deep neural network (DNN) model. We first map input tokens into a share-space word embedding, and then feed them into a biLSTM encoder to obtain context information; this content then propagates to the downstream tasks, with CRF used for NER prediction, and an MLP classifier used for IC prediction. We call this a self-trained multilingual (STM) model. Figure \ref{fig:1} shows the model architecture.

In the transfer learning experiments, we examined the following approaches:
\begin{enumerate}
	\item Trained model on high-resource language(s), then transferred both encoder and decoder to the new multi-lingual model with fine-tuning (EncDecTL models).
	\item Trained model on high-resource language(s), then transferred only encoder with fixed parameters to the new multi-lingual model (EncTL models).
	\item Trained model on high-resource language(s), then transferred only encoder with variable learning rate, that we gradually unfreeze embeddings with training steps during fine-tuning (EncVLRTL models).
\end{enumerate}
Figure 2(a) shows the architecture for approach 1, and Figure 2(b) shows the architecture for approaches 2 and 3.

\begin{figure}[ht]
	\centering
	\begin{subfigure}[b]{0.43\textwidth}
		\includegraphics[width=\textwidth]{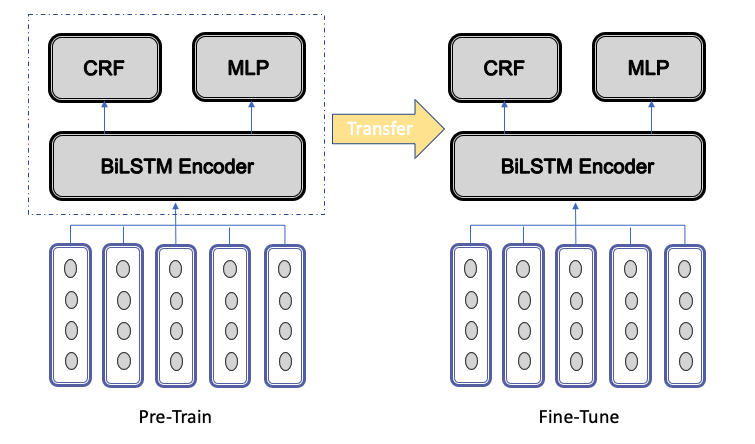}
		\caption{Both encoder and decoder are transferred}
		\label{fig:2a}
	\end{subfigure}%
	\quad
	\begin{subfigure}[b]{0.44\textwidth}
		\centering
		\includegraphics[width=\textwidth]{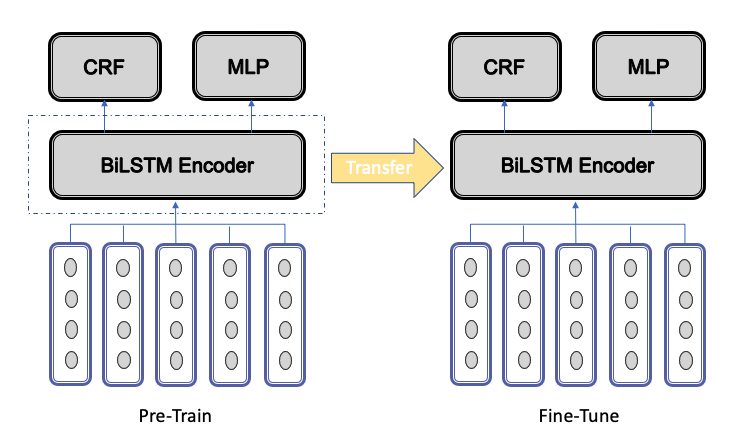}
		\caption{Only encoder is transferred}
		\label{fig:2b}
	\end{subfigure}
	\caption{Transfer learning model architectures}\label{fig:2}
\end{figure}

We limit our architecture to a small-sized DNN model to avoid memory and latency issues in production environment:
\begin{itemize}
	\item \textbf{Embedding}: Concatenation of share-spaced word embedding and character embedding trained on a 3-filter convolutional neural network (CNN). All embeddings for DNN models are initialized with fastText supervised multilingual word embeddings, aligned in a single vector space (Conneau et al., 2017). Aligned multilingual word embeddings kept fixed in pre-training model, and trained during fine-tuning.
	\item \textbf{Encoder}: 2-layer biLSTM with 512 dimensions in each hidden layer.
	\item \textbf{Decoders}: Both the MLP classifier used for IC task and the CRF sequence labeler used for the NER task have 512 dimensions in each block's hidden layer, and GELU activation function. We also used dropout to prevent overfitting. All models were trained for 160 epochs with early stopping (experiments were attempted to increase the number of epochs, yet this did not show much change in model performance).
	\item \textbf{Loss function}: A combination of loss from both tasks of IC and NER: $L_{total} = \alpha L_{ic} + \beta L_{ner}$ , where $\alpha$ and $\beta$ are the associated weight of each loss contribution. In our experiments, we have fixed both $\alpha$ and $\beta$ to 1.
\end{itemize}

\section{Experiments and discussion}
\subsection{Data}
We train our models using data collected for four languages, including three relatively closely related language: UK English, Spanish, and Italian. In additon to these three similar languages, we also collected a dataset of Hindi, a language which is lexically and grammatically different from the other three. All utterances represent users' requests and are annotated with corresponding DC, IC, and NER tags. Summary statistics of the data set is shown in Table 1. We limit our experiments to the seven NLU domains: Communication, General Media, Home Automation, Music, Notifications, Shopping, and Video. Each training set is further divided into training and validation split in ratio 9:1.

\begin{table*}[ht]
	\centering
	\label{tab:summary-table}
	\centering
	\begin{small}
		\footnotesize
		\begin{tabular}{lccc}
			\hline
			\textbf{Language} & \textbf{Number of utterances in train/test split} & \textbf{Intent types} & \textbf{Slot types} \\ \hline
			English & 2,121,583 / 117,546 & 316 & 282 \\ \hline
			Spanish & 2,927,850 / 144,156 & 365 & 311 \\ \hline
			Italian & 2,435,459 / 107,298 & 379 & 324 \\ \hline
			Hindi & 370,465 / 69,024 & 302 & 267 \\ \hline
		\end{tabular}
	\end{small}
	\caption{Summary statistics of the data set}
\end{table*}

\subsection{Results and discussion}
Following \cite{schuster-etal-2019-cross-lingual}, we evaluate our models according to four metrics: domain accuracy for the DC task, intent accuracy for the IC task, micro-averaged slot F1 for NER prediction, and frame accuracy which is the relative number of utterances for which the domain, intent, and all slots are correctly identified. Relative performance for all model architectures evaluated on each language-specific test set is shown in Figure 3. Metrics are averaged across three model runs for English, Spanish and Italian test results, whereas two model runs were performed for Hindi; detailed numbers are provided in Appendix A. The baseline models used for the relative metric performance are monolingual models with a MaxEnt classifier for IC and a CRF model for NER (\textit{Mono MaxEnt}) trained on data set which consists of utterances in the target language only. For example, in the case of Spanish, the relative performance for all models is calculated against the Mono MaxEnt model trained on Spanish data only.

\begin{figure*}[ht]
	\centering
	\begin{subfigure}[b]{0.43\textwidth}
		\includegraphics[width=\textwidth]{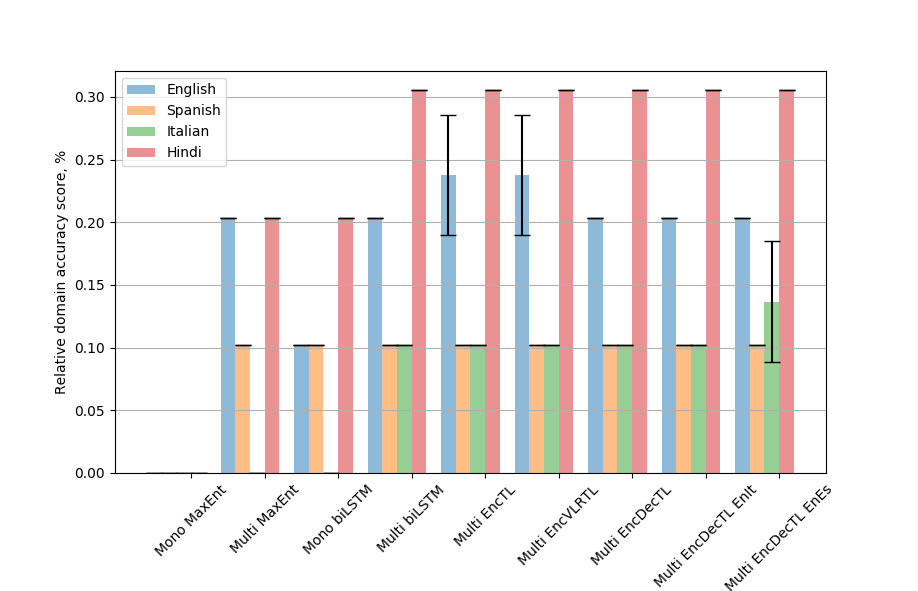}
		\caption{Relative domain accuracy score}
		\label{fig:3a}
	\end{subfigure}%
	\quad
	\begin{subfigure}[b]{0.43\textwidth}
		\centering
		\includegraphics[width=\textwidth]{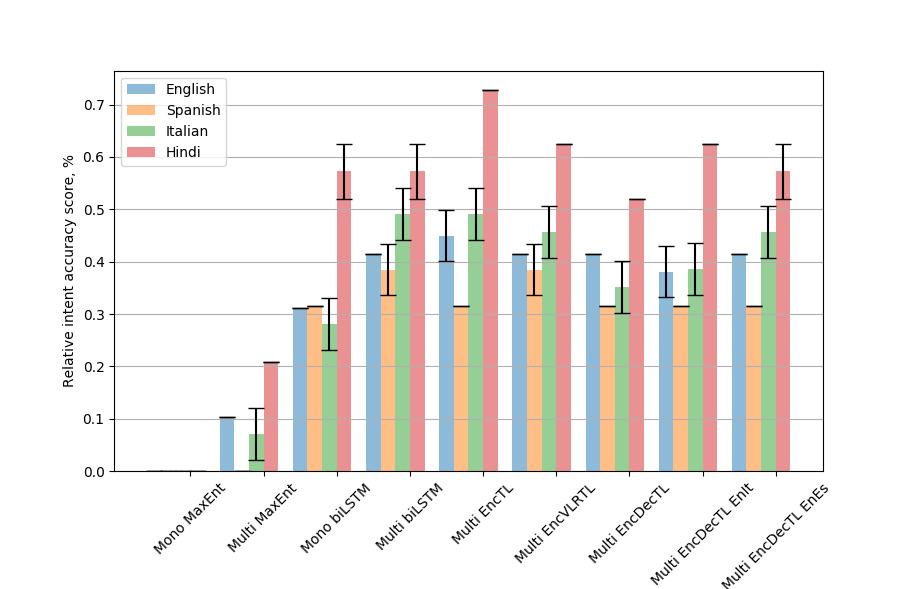}
		\caption{Relative intent accuracy score}
		\label{fig:3b}
	\end{subfigure}
	\quad
	\begin{subfigure}[b]{0.43\textwidth}
		\centering
		\includegraphics[width=\textwidth]{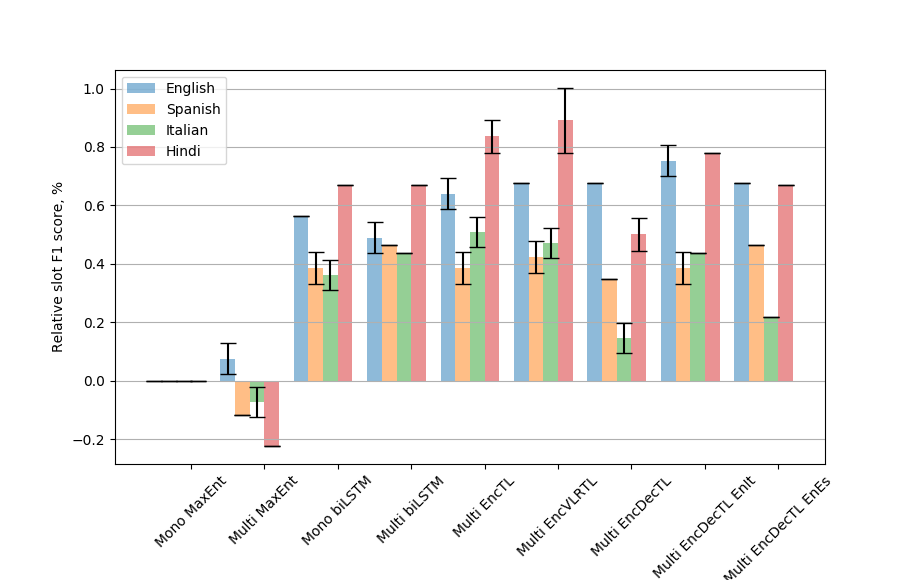}
		\caption{Relative slot F1 score}
		\label{fig:3c}
	\end{subfigure}
	\quad
	\begin{subfigure}[b]{0.43\textwidth}
		\centering
		\includegraphics[width=\textwidth]{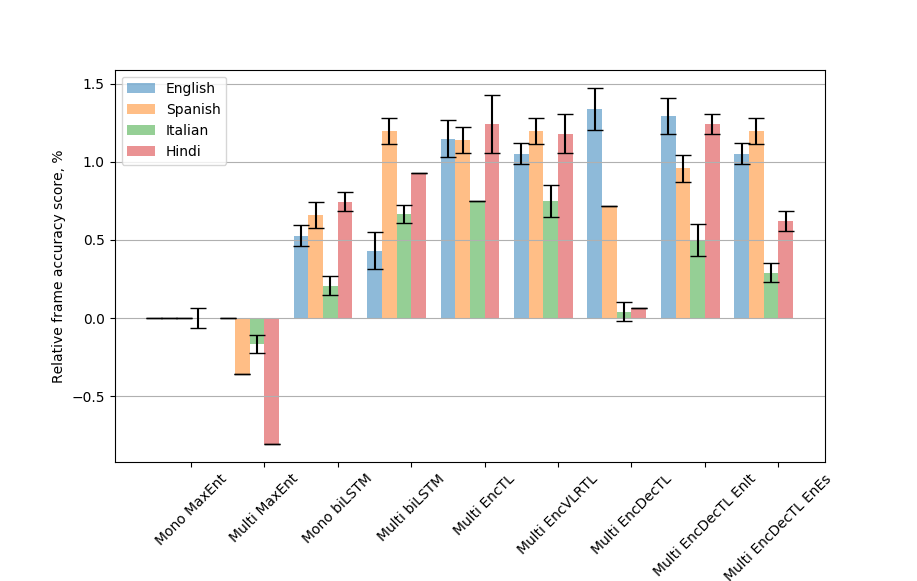}
		\caption{Relative frame accuracy score}
		\label{fig:3d}
	\end{subfigure}
	\caption{Percentage change in model performance for mono- and multilingual models with respect to the monolingual MaxEnt model. All models for transfer learning were pre-trained on English data. In addition, two encoder-decoder transfer learning models were built on the mix of English and Italian (EnIt) and English and Spanish (EnEs) models}\label{fig:3}
\end{figure*}

All performance metrics show similar pattern: multilingual DNN models usually perform better than monolingual models, benefitting from biLSTM setup and encoder transfer learning. The only exception is domain accuracy score, where we observe only slight variations for mono- and multilingual models. The reason is because the DC model in all experiments is governed by MaxEnt model and is not affected by the changing DNN setup, but only training data mixing. Non-English models benefit from encoder transferred from pre-trained mono- and multilingual models, whereas additional decoder transfer slightly degrades model performance, but still beats the baseline. Mixing languages in pre-training phase (Table 2 only slightly improves target model performance). For English, EncDecTL model with transfer learning performs the best across all metrics, with EncTL model showing similar performance; while for Spanish and Italian, EncVLRTL pretrained on English is the best model among the pool (see Appendix A for detailed pairwise model comparison). Overall the best average performance across the languages is observed in DNN models with encoder transfer learning, which has about 1\% improvement in frame accuracy. Interestingly, more improvement ($>1\%$) in frame accuracy is observed in applying transfer learning on Hindi; a lexically and grammatically different language than the family of the pretrained languages. This improvement may be due to the fact that the baseline model for Hindi language was trained on a low-resource dataset, thus the encoder and decoders can learn more information from the added data, whereas encoder and decoder learning from a similar language have already been near-to-optimal, and the added data do not contribute more information gain in training. Default DNN models without transfer learning are the next best performing models (Table 3). These model results demonstrate that it is possible not only to create a single model that will serve multiple languages simultaneously, but also to expect some performance improvements.
\begin{table*}[ht]
	\centering
	\label{tab:multi-result}
	\centering
	\begin{small}
		\footnotesize
		\setlength\extrarowheight{1.5pt}
		\begin{tabular}{p{1.2cm}p{2.4cm}ccccc}
			\hline
			\textbf{Target language} & \textbf{Pre-training language} & \textbf{SEMER} & \textbf{Domain accuracy} & \textbf{Intent accuracy} & \textbf{Slot F1} & \textbf{Frame accuracy} \\ \hline
			\multirow{2}{2cm}{Italian} & English           			& 1.004 & 1.000 & 0.999 & 0.997 & 0.994 \\ \cline{2-7}
			{} 										& English \& Spanish & 1.002 & 1.000 & 1.000 & 0.998 & 0.996 \\ \hline
			\multirow{2}{2cm}{Spanish} & English           		& 1.000 & 1.000 & 0.999 & 0.996 & 0.996 \\ \cline{2-7}
			{} 											& English \& Italian & 1.000 & 1.000 & 0.999 & 0.999 & 0.998 \\ \hline
		\end{tabular}
	\end{small}
	\caption{\textit{Multi-lingual biLSTM} model performance for the targeted languages with monolingual and multilingual encoder-decoder transfer learning on both Italian and Spanish test data. (Values are normalized based on \textit{Multi-lingual biLSTM}. See Appendix A for more details.)}
\end{table*}
\begin{table*}[ht]
	\centering
	\label{tab:average-performance}
	\centering
	\begin{small}
		\footnotesize
		\begin{tabular}{m{3.5cm}>{\centering\arraybackslash}m{3.5cm}>{\centering\arraybackslash}m{3.5cm}>{\centering\arraybackslash}m{3.5cm}}
			\hline
			\textbf{Model} & \textbf{Average intent accuracy improvement$(\%)$} & \textbf{Average slot F1 improvement$(\%)$} & \textbf{Average frame accuracy improvement$(\%)$} \\ \hline
			DNN without transfer learning & 0.45 & 0.45 & 0.71 \\ \hline
			DNN with encoder transfer learning & 0.42 & 0.53 & 0.99 \\ \hline
			DNN with encoder transfer learning and variable learning rate & 0.42 & 0.53 & 1.00 \\ \hline
		\end{tabular}
	\end{small}
	\caption{Average multilingual model performance change across English, Spanish and Italian test sets for different DNN setups.}
\end{table*}

Deep diving into EncDecTL errors for non-English languages, we found that the failures occur due to the differences in intent and slot label distributions in training data across different languages. When intents or slot labels from target language are missing in the pre-trained model, these entities are not represented in decoder’s vocabulary. In our setup, the decoder’s vocabulary is being transferred along with weights matrices, so the decoder fails to learn and predict new labels simply because they are missing from the list of the possible outputs. Since differences in slot labeling are more common than in intent labelling, we observe higher degradations for slot F1 and frame accuracy scores than for intent accuracy. Encoder, on the other hand, operates with aligned multilingual word embeddings only, and is not affected by the differences in annotations. This label/intent missing issue could be mitigated by the expanding decoder’s vocabulary in the fine-tuning model with intent and slot labels specific to the target language.

Finally, to understand how the DNN models affect latency, we measure average time taken by the statistical model to evaluate performance on the full multilingual test set over three runs. We found that the latency between the models is comparable, the DDN models even have a slightly lower relative average runtime of 0.95 versus 1.05 for MaxEnt models (detailed can be found in Appendix B for each model architecture).

\section{Conclusions}
In this paper, we propose a framework for building general multilingual NLU models, which can be used across different marketplaces and languages.

To choose the model with the best performance, we use language-specific test sets to evaluate the candidate models and their corresponding baseline models (e.g. English biLSTM-CRF model vs. monolingual English MaxEnt-CRF model) along four metrics, domain accuracy, intent accuracy, slot F1, and frame accuracy. The models which win in most of the evaluation metrics are the final picks. We find that models built from a simple multi-task biLSTM-CRF model setup are comparable to standard production models in terms of latency constraints required for on-the-fly voice assistant conversational models.

We observe performance improvements in all models with the introduction of transfer learning. Encoder transfer produced the greatest improvements whereas the transfer of the decoder did not bring much change when compared to the baseline model performance, except when tested on an English test set, when the transfer learning is performed from the model trained on English data. This is due to the fact that the target non-English language contains slots or intents which are not included in the pre-trained model, thus the decoder fails to predict correct classes simply because they are missing in the vocabulary. To mitigate this effect, a decoder with default initialization gives better performance because it now can embrace all available slots and intents in the target language realm.

Furthermore, we find that a model pre-trained in a multilingual setup performs better than the one trained on a monolingual data set. This confirms that a multilingual model built based on lexically and orthographically similar languages may provide more beneficial context information to any similar target language. Experimental result on Hindi show that such a multilingual model can work even for non-alike languages with the same or better performance improvement. This confirms that a common multilingual model can be used to support multiple language with better results than a set of monolingual models.

With a single general multilingual NLU model, bootstrapping new languages can be faster as we can use cross-lingual contextual information from all existing high-resource languages. At the same time, maintaining only one model requires much less effort in terms of regular model updates.

\bibliographystyle{acl}
\bibliography{coling2020}

\pagebreak
\begin{appendices}
	\section{Multilingual test results on language-specific datasets}
	Three model runs are performed on English, Spanish and Italian, while two model runs are performed on Hindi. All values are normalized based on \textit{Multi-lingual biLSTM}, the 2-layer biLSTM CRF cross-lingual model.
	\begin{figure}[ht]
		\centering
		\includegraphics[width=1.2\linewidth, angle=270]{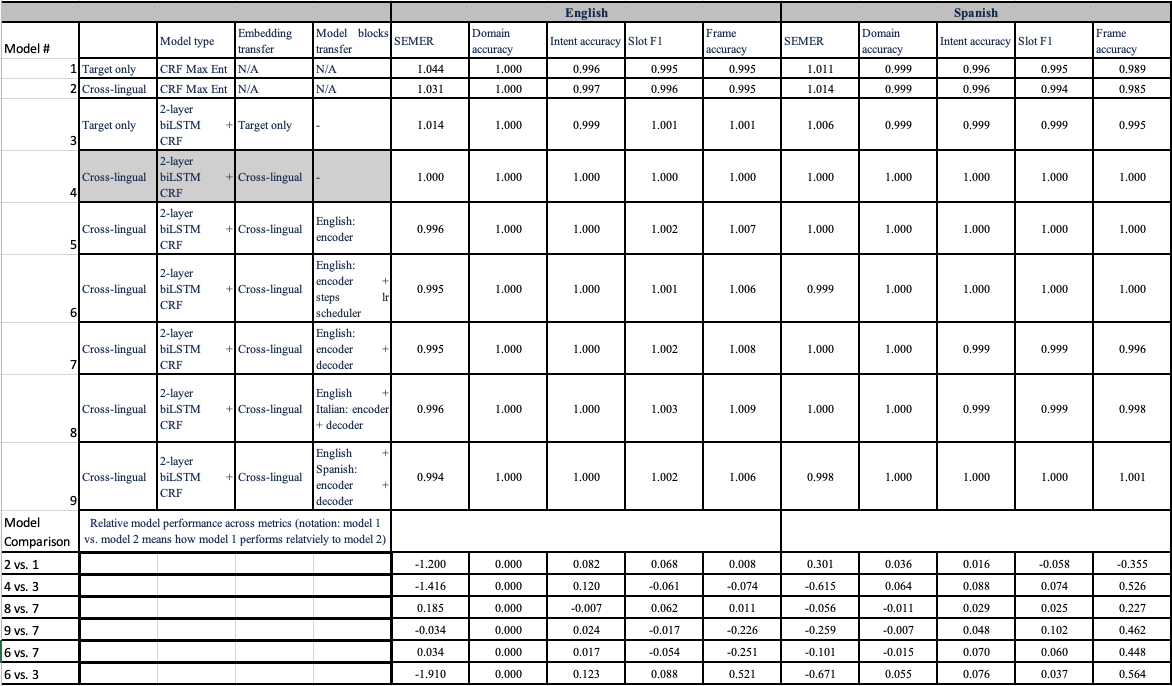}
	\end{figure}
	\begin{figure}[ht]
		\centering
		\includegraphics[width=0.8\linewidth, angle=270]{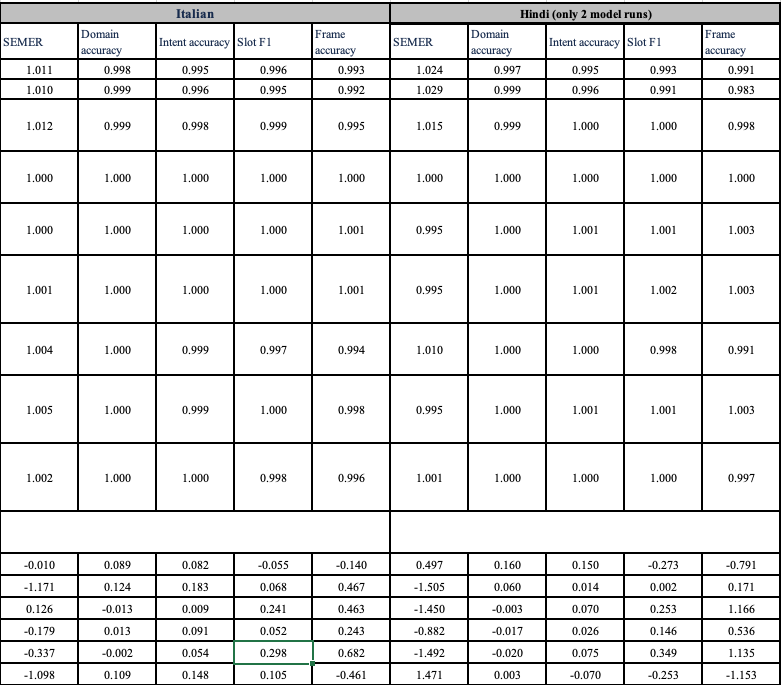}
	\end{figure}

	\pagebreak
	\section{Average relative evaluation runtime for each model on the full test set}
	All values are normalized based on \textit{Multi-lingual biLSTM}, the 2-layer biLSTM CRF cross-lingual model.
	\begin{table*}[ht]
		\centering
		\resizebox{\textwidth}{!}{%
			\begin{tabular}{|l|l|l|l|l|c|}
				\hline
				\textbf{Model \#} &
				\textbf{Training set} &
				\textbf{Model type} &
				\textbf{Embedding transfer} &
				\textbf{Model blocks transfer} &
				\multicolumn{1}{l|}{\textbf{Runtime (rel)}} \\ \hline
				\textbf{1} & Target only   & CRF Max Ent           & N/A           & N/A                                   & 1.00 \\ \hline
				\textbf{2} & Cross-lingual & CRF Max Ent           & N/A           & N/A                                   & 1.05 \\ \hline
				\textbf{3} & Target only   & 2-layer biLSRTM + CRF & Target only   & -                                     & 0.97 \\ \hline
				\textbf{4} & Cross-lingual & 2-layer biLSRTM + CRF & Cross-lingual & -                                     & 1.00 \\ \hline
				\textbf{5} & Cross-lingual & 2-layer biLSRTM + CRF & Cross-lingual & English: encoder                      & 0.97 \\ \hline
				\textbf{6} & Cross-lingual & 2-layer biLSRTM + CRF & Cross-lingual & English: encoder + steps lr scheduler & 0.95 \\ \hline
				\textbf{7} & Cross-lingual & 2-layer biLSRTM + CRF & Cross-lingual & English: encoder + decoder            & 0.98 \\ \hline
				\textbf{8} & Cross-lingual & 2-layer biLSRTM + CRF & Cross-lingual & English + Italian: encoder + decoder  & 1.02 \\ \hline
				\textbf{9} & Cross-lingual & 2-layer biLSRTM + CRF & Cross-lingual & English + Spanish: encoder + decoder  & 1.02 \\ \hline
			\end{tabular}%
		}
	\end{table*}

\end{appendices}
\end{document}